\documentclass[runningheads]{llncs}
\usepackage[T1]{fontenc}
\usepackage{orcidlink}
\usepackage{graphicx} 
\usepackage{amsmath}
\usepackage{amssymb} 
\usepackage{booktabs}
\usepackage{microtype}
\usepackage{subcaption}
\usepackage{xcolor}
\usepackage{colortbl}

\usepackage{float}

\definecolor{mygreen}{rgb}{0.0, 0.5, 0.0} 
\newcommand{\cpos}[1]{\cellcolor{green!25}#1}
\newcommand{\cneg}[1]{\cellcolor{red!25}#1}

\usepackage{hyperref}
\usepackage{color}

\usepackage[backend=biber,style=numeric-comp,giveninits=true,isbn=false,url=false, sorting=none,maxbibnames=2]{biblatex}
\title{Analyzing Shapley Additive Explanations to Understand Anomaly Detection Algorithm Behaviors and Their Complementarity}

\titlerunning{SHAP-based Complementarity in UAD}

\author{Jordan Levy\inst{1,2}\orcidlink{0009-0002-4337-2608} \and
Paul Saves\inst{1}\orcidlink{0000-0001-5889-2302} \and
Moncef Garouani\inst{1}\orcidlink{0000-0003-2528-441X}
Nicolas Verstaevel\inst{1}\orcidlink{0000-0002-7879-6681} \and
Benoit Gaudou\inst{1}\orcidlink{0000-0002-9005-3004}}

\authorrunning{J. Levy, P. Saves, M. Garouani, N. Verstaevel and B. Gaudou}

\institute{Institut de Recherche en Informatique de Toulouse, Université Toulouse Capitole, Toulouse, France \\\email{jordan.levy@irit.fr, paul.saves@irit.fr, moncef.garouani@irit.fr, nicolas.verstaevel@irit.fr, benoit.gaudou@irit.fr}\and
TwinswHeel, Fontanes, France\footnote{IDA 2026 Frontier Prize, Best paper, Springer Nature.}}

\begin{document}

\maketitle

\begin{abstract}
	Unsupervised anomaly detection is a challenging problem due to the diversity of data distributions and the lack of labels. Ensemble methods are often adopted
	to mitigate these challenges by combining multiple detectors, which can reduce individual biases and increase robustness. Yet building an ensemble that is
	genuinely complementary remains challenging, since many detectors rely on similar decision cues and end up producing redundant anomaly scores. As a result,
	the potential of ensemble learning is often limited by the difficulty of identifying models that truly capture different types of irregularities. To address
	this, we propose a methodology for characterizing anomaly detectors through their decision mechanisms. Using SHapley Additive exPlanations, we quantify how
	each model attributes importance to input features, and we use these attribution profiles to measure similarity between detectors. We show that detectors
	with similar explanations tend to produce correlated anomaly scores and identify largely overlapping anomalies. Conversely, explanation divergence reliably
	indicates complementary detection behavior. Our results demonstrate that explanation-driven metrics offer a different criterion than raw outputs
	for selecting models in an ensemble. However, we also demonstrate that diversity alone is insufficient; high individual model performance remains a prerequisite for effective ensembles.
	By explicitly targeting explanation diversity while maintaining model quality, we are able to construct ensembles that are more diverse, more complementary,
	and ultimately more effective for unsupervised anomaly detection.

\end{abstract}

\section{Introduction}
Anomaly detection is a challenging problem, primarily because of the intrinsic nature of anomalies. In fact, anomalies are defined as deviations from what is
considered to be normal behavior~\cite{chandola2009anomaly}. Therefore, depending on how this normality is defined, one algorithm or another may be more
adapted for good detection. For instance, some methods may define normality in geometric terms and use distance-based rules to spot outliers, while some others
may leverage probabilistic models that flag rare or low-probability instances as anomalies~\cite{ruff2021unifying}.

In many real-world settings, obtaining labeled anomalies is expensive or infeasible because anomalous events are rare, costly to produce, or hazardous to
induce (\textit{e.g.}, in the nuclear industry where detection must operate without waiting for faults to occur)~\cite{oh2022anomalies}. Consequently,
semi-supervised and unsupervised approaches are often preferred. In purely unsupervised anomaly detection (UAD), practitioners typically rely on machine
learning methods because the “normal” behavior cannot be specified exhaustively by hand; therefore, algorithm designers must adopt implicit assumptions about
normality (for instance, geometric/distance-based or probabilistic models), and no single assumption is guaranteed to be appropriate across all
applications~\cite{aggarwal2016outlier}. Furthermore, in a fully unsupervised setting, anomaly detection algorithms cannot be directly benchmarked due to the
lack of ground truth labels and, as stated in the \textit{no free lunch theorem}, no single detector consistently outperforms others across all
datasets~\cite{wolpert2002no}. This theorem is also valid for anomaly detection~\cite{han2022adbench}. Nevertheless, the choice of algorithm remains critical,
as an inappropriate selection may lead to poor performance~\cite{aggarwal2016outlier}.

A common strategy to mitigate this issue is to combine multiple detectors in an ensemble framework, leveraging the strengths of each model and its diverse
assumptions about what constitutes a normal behavior. Ensemble learning has been widely adopted for this reason and has demonstrated strong empirical
performance across various anomaly detection scenarios~\cite{aggarwal2016outlier}. The advantage of ensemble methods in anomaly detection lies in their ability
to combine models that rely on different underlying assumptions about normality. By aggregating detectors with diverse inductive biases, ensembles can capture
a wider variety of anomaly types. This diversity often translates into improved detection coverage~\cite{schubert2012evaluation}.

A key challenge in ensemble learning is the selection of appropriate base detectors, as benchmarking with ground truth labels is typically impossible in UAD. While a robust ensemble theoretically requires models that are both diverse and performant~\cite{schubert2012evaluation}, identifying such complementary behaviors remains an open issue~\cite{ma2023need}.

To address this challenge, we propose a novel methodology that characterizes the behavior of anomaly detectors using SHapley Additive exPlanations (SHAP)~\cite{lundberg2017unified}. Unlike approaches relying solely on outputs, we focus on understanding the internal decision mechanisms of the algorithms. Our study reveals that detectors with similar explanation patterns tend to produce redundant anomaly scores, whereas explanation divergence is a strong indicator of complementarity. Consequently, we demonstrate that selecting detectors based on their explanation behavior leads to improved ensemble performance. This leads to three major contributions:

\begin{itemize}
	\item An analysis of UAD algorithms based on their SHAP explanations, which demonstrates that their behaviors are correlated to their output similarity.
	\item A comparison between SHAP explanations and model output similarities for selecting diverse anomaly detector models.
	\item An empirical study quantifying the relative importance of diversity versus individual performance, demonstrating that model quality remains a critical prerequisite.
\end{itemize}
This article is structured as follows: Section~\hyperref[sec:rltdwork]{2} presents the related works. The methodology of our approach is described in Section~\hyperref[sec:methodo]{3}. Experimental results are introduced in Section~\hyperref[sec:exp]{4}. Finally, we conclude the article in Section~\hyperref[sec:conclusion]{5}.

\section{Related Work}
\label{sec:rltdwork}

\subsection{Unsupervised Anomaly Detection Algorithms}

UAD is a widely studied problem~\cite{ruff2021unifying} with a large number of algorithms, each founded on different assumptions regarding the nature of
anomalies. Libraries like PyOD~\cite{zhao2019pyod} have been introduced to standardize the usage of these models. Among these, distance-based methods (KNN,
LOF, CBLOF) characterize anomalies based on proximity. Conversely, reconstruction-based algorithms (AutoEncoder, PCA) detect anomalies by learning to
reconstruct normal data patterns. Other approaches, such as HBOS, ECOD, and COPOD, leverage probabilistic assumptions, while OCSVM and DeepSVDD utilize
one-class classification. Finally, algorithms like Isolation Forest and LODA rely on random partitioning and space projection, respectively. Each of these
algorithms has a specific hypothesis regarding the nature of anomalies; however, empirical evidence suggests that no single detector consistently outperforms
all others~\cite{han2022adbench}. Consequently, employing an ensemble of multiple models serves as a robust strategy, effectively leveraging diverse underlying
hypotheses to improve detection capabilities~\cite{aggarwal2016outlier}.

\subsection{Model Selection for Unsupervised Anomaly Detection}

Selecting an appropriate model for an UAD task is commonly known as {Unsupervised Outlier Model Selection (UOMS)}. UOMS is a particularly challenging problem
that has gained increasing importance as the number of available models, comprising various algorithmic families and hyperparameter configurations, keeps
expanding as new methods appear in the literature~\cite{ruff2021unifying}.

To that end, a common strategy is to estimate model quality directly from unlabeled data in an unsupervised way, to select the best models. These approaches
fall into two main categories. \emph{Stand-alone} methods~\cite{goix2016evaluate,marques2020internal} compute an unsupervised score for each
detector independently, while \emph{consensus-based} methods assess detectors by measuring agreement within a pool of models and selecting those that best
conform to the group~\cite{duan2019unsupervised,lin2020infogan}. Empirical evidence indicates that stand-alone evaluation criteria often fail to deliver
consistent or reliable results in practice. In contrast, consensus-based approaches, despite their higher computational cost, have emerged as a more effective
and promising strategy for unsupervised outlier model selection~\cite{ma2023need}.

There is limited work in the scientific literature on UOMS for ensemble learning. In~\cite{schubert2012evaluation}, the authors show that diversity in
algorithm hypotheses tends to give better results. To characterize diversity, the authors use a weighted Pearson correlation between the anomaly scores of the
models. However, they also highlight that diversity is important, but the chosen algorithms should already have good performance on the dataset to have a
performant ensemble model. From the previous assumption, the authors in~\cite{rayana2016less} introduced SELECT, which, instead of trying to select models
according to their diversity, selects models according to their performances from a pseudo ground-truth label created from anomaly scores.
More recently, in~\cite{campos2021unsupervised}, the authors perform anomaly detection in time series using multiple convolutional autoencoders as an ensemble and optimize diversity explicitly by incorporating a diversity metric based on the dissimilarity of the reconstruction outputs. They also demonstrate that this diversity-driven strategy improves detection performance.

While it has been shown that models output like anomaly scores can reflect diversity, in this work, we investigate a consensus method based on SHAP values, which characterize the
behavior of different anomaly detection models across different datasets. Furthermore, in our work, we quantify the impact of diversity and propose a methodology to
model the relationship between mean individual performance, diversity, and the resulting ensemble gain.

\subsection{Explainability}

Explainable Artificial Intelligence (XAI) seeks to make complex models transparent by producing human-interpretable accounts of how inputs, model structure,
and uncertainty produce particular outputs~\cite{madsen2024interpretability}. Among local, model-agnostic explanation methods, SHapley Additive exPlanations
(SHAP) are widely used because they provide axiomatic, instance-level feature attributions grounded in the Shapley value principle~\cite{lundberg2017unified}.

Beyond interpretability, SHAP representations have been exploited to compare models and to form informative embeddings: recent works use aggregated SHAP
attributions to identify model families or to serve as features for downstream learning tasks~\cite{saves2025surrogatemodelingexplainableartificial,garouani2025xstacking}. To our knowledge, leveraging SHAP for model comparison and selection has tot been systematically investigated in the UAD setting, where the lack of labels makes model interpretability and selection particularly challenging.

\section{Methodology}
\label{sec:methodo}
\subsection{Problem Setting}
We consider the problem of UAD, where the goal is to identify anomalous samples within a dataset $\mathcal{D} = \{x_1, \dots, x_n\}$ containing $n$ instances
in a $d$-dimensional feature space. We assume a set of $m$ anomaly detection models $\mathcal{M} = \{ M_1, M_2, \dots, M_m \}$. For a given dataset, every
model $M_i$ produces an anomaly score output vector $s_i = M_i(\mathcal{D}) \in \mathbb{R}^n$ where $s_i^{(k)}$ is the anomaly score of the $k^\text{th}$ input
point according to the $i^\text{th}$ model. Our objective is to analyze the behavior of these models and to identify groups of models that share similar
interpretative structures.

\subsection{Model Similarity from Explanations}
For each model $M_i$, we compute its SHAP explanation matrix \( Sh_i \in \mathbb{R}^{n \times d},\) where $Sh_i^{(k)} \in \mathbb{R}^{d}$ is the vector
representing the contribution of every feature to the anomaly score of instance $x_k$.

We define the behavior similarity between two models $M_i$ and $M_j$ as the average per-instance Pearson correlation between their SHAP vectors:
\begin{equation*}
	\rho_{ij}^{\text{PS}} =
	\frac{1}{n} \sum_{k=1}^n
	\text{corr}\big(Sh_i^{(k)}, Sh_j^{(k)}\big).
\end{equation*}

To capture ranking consistency between feature importances rather than raw magnitudes, we also compute a similarity based on the Normalized Discounted
Cumulative Gain (NDCG):
\begin{equation*}
	\rho_{ij}^{\text{NDCG}} =
	\frac{1}{2n} \sum_{k=1}^n
	\Big(
	\operatorname{NDCG}\big(|Sh_i^{(k)}|, |Sh_j^{(k)}|\big)
	+
	\operatorname{NDCG}\big(|Sh_j^{(k)}|, |Sh_i^{(k)}|\big)
	\Big),
\end{equation*}
where NDCG evaluates the agreement in the ranked importance of features between two detectors. If both models assign high SHAP importance to the same features, their NDCG value will be close to 1, indicating consistent interpretative behavior~\cite{burges2005learning}.

\subsection{Linking Similarities to Detection Performances}

To compare the similarity of models with performances, we introduce two additional matrices: the matrix of score correlations and the matrix of Jaccard
similarities.

We define the similarity between two models \( M_i \) and \( M_j \) based on the correlation of their anomaly scores. Specifically, we compute the average
per-instance Pearson correlation between their score vectors as:
\begin{equation*}
	\rho_{ij}^{\text{Score}} =
	\frac{1}{n} \sum_{k=1}^{n}
	\mathrm{corr}\big(s_i^{(k)}, s_j^{(k)}\big).
\end{equation*}
This results in a symmetric matrix capturing the pairwise similarity between models in terms of their scoring behavior.

To compare model outputs directly, we define $A_i$ as the set of instances identified as anomalies by model $i$ (where the anomaly score exceeds a threshold
$\tau_i$). The Jaccard similarity between two models $i$ and $j$ is defined as:

\begin{equation*}
	J_{ij} = \frac{|A_i \cap A_j|}{|A_i \cup A_j|}.
\end{equation*}

This metric measures the overlap between models: a value of 1 implies identical predictions, while 0 implies disjoint sets of detected anomalies.

To quantify the relationship between the different measures, we first transform each similarity matrix $P \in \{\rho^{PS}, \allowbreak \rho^{NDCG}, \allowbreak
	\rho^{Scores}, \allowbreak J\}$ into a dissimilarity matrix $D^P_{\mathcal{D}}$. This is computed as $D^P_{\mathcal{D}} = 1 - P$, representing the distance
between two detectors for a given dataset $\mathcal{D}$. We then apply the Mantel test~\cite{mantel1967detection} to determine whether two given dissimilarity
matrices are statistically correlated. The Mantel coefficient ($r_M$) is calculated as the Pearson
correlation between the upper triangular elements of the matrices, with statistical significance established via permutation testing. In this context, a
significant positive $r_M$ implies that anomaly detectors sharing similar SHAP-based explanations also exhibit similar anomaly scores.

\section{Experiments}
\label{sec:exp}

We conducted experiments to answer the following: (1) Does explanation similarity imply prediction similarity? (2) Do SHAP-based metrics outperform raw outputs
in quantifying diversity? and (3) How significantly does this diversity impact the accuracy and robustness of the resulting ensemble?

\subsection{Experimental Setup}
In the experiments, 14 UAD algorithms were used. Specifically, COF, KNN, LOF, IForest, PCA, CBLOF, LODA, HBOS, MCD, OCSVM, DAGMM, DeepSVDD, COPOD, and ECOD
were employed. For the implementation of each model, the Python library PyOD~\cite{zhao2019pyod} was used. The hyperparameters defined by the authors of
each model were retained, and no indication of the percentage of anomaly was provided for any dataset, keeping the threshold to its default value. Regarding the explainability component, we used the model-agnostic Kernel SHAP implementation of the Python package \texttt{shap}. This choice was necessary to ensure a unified explanation framework across our diverse algorithms. Furthermore, to ensure approximation stability, the background dataset was summarized using a k-means clustering with $k=50$ centroids.

To further strengthen the robustness of the results, each dataset was repeatedly split into a training set and a test set five times, with the training set representing 80\% of the data. The random seed for each split was set equal to the iteration index to ensure reproducibility. The source code used in the experiments is available on GitHub\footnote{\url{https://github.com/jordanlv/Analyzing-SHAP-UOMS}}

\subsection{Datasets Considered}
Due to the high computational cost associated with SHAP, the dataset selection was restricted to the 50\% smallest datasets available in
ADBench~\cite{han2022adbench}. Additionally, datasets containing more than 20 features were removed to maintain computational feasibility. This initial
filtering process resulted in a set of 16 datasets from various applications: annthyroid (AN), breastw (BR), glass (GL), Hepatitis (HE), Lymphography (LY),
mammography (MA), PageBlocks (PB), Pima (PI), Stamps (ST), thyroid (TH), vertebral (VE), vowels (VO), WBC (WB), Wilt (WL), wine (WN) and yeast (YE). Datasets characteristics can be found in Table~\ref{tab:datasets} in Appendix~\ref{sec:appendixa}.

\subsection{Correlation Between Similarity Matrices}
\begin{figure}[htb!]
	\centering
	\includegraphics[width=\textwidth]{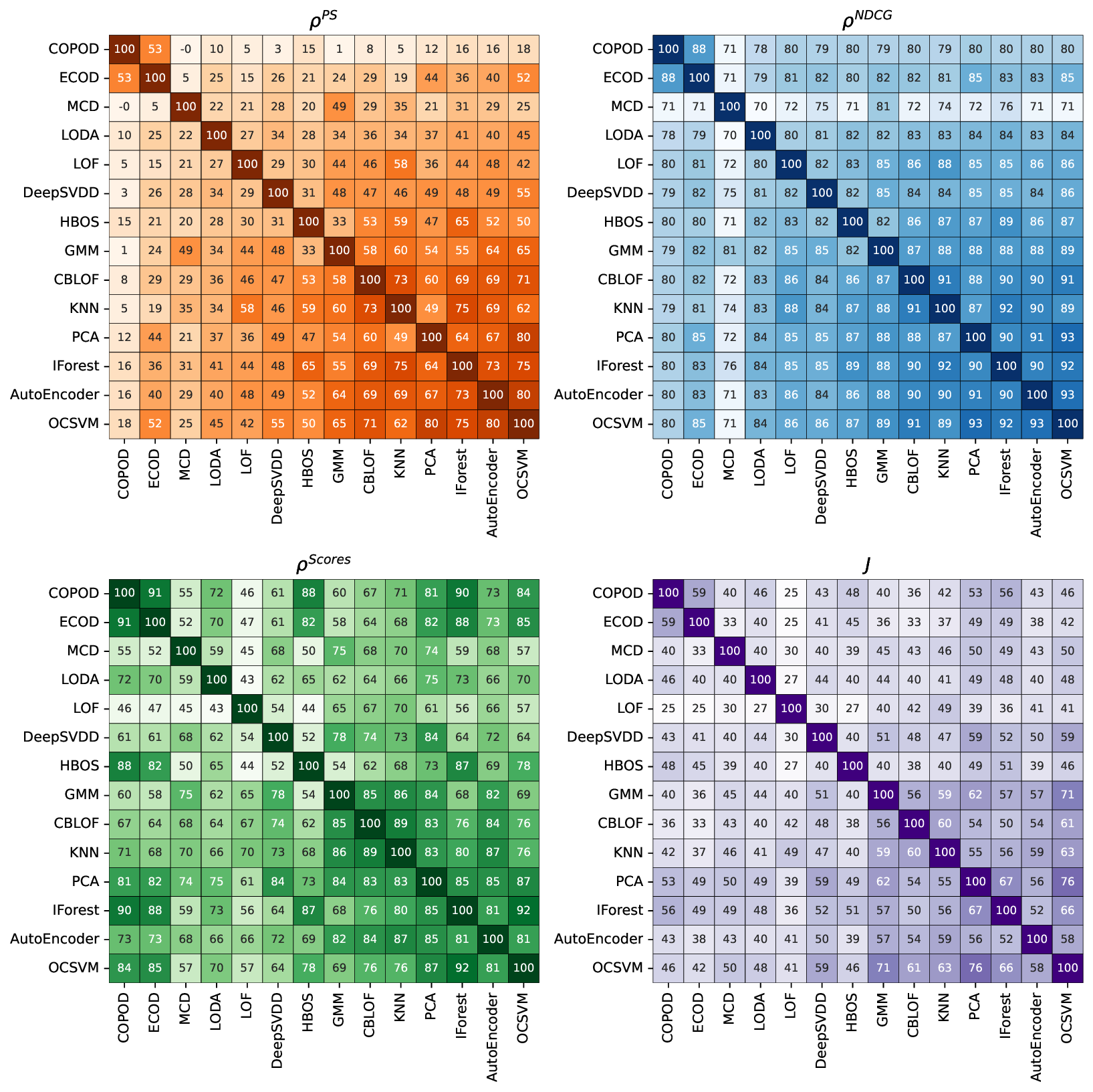}
	\caption{Mean similarity between models across all datasets. Layout: Top-Left: correlations of SHAP values; Top-Right: NDCG of SHAP values; Bottom-Left: correlations of anomaly scores; and Bottom-Right: Jaccard similarities.}
	\label{fig:matrices}
\end{figure}
We computed the four similarity matrices $\rho^{PS}$ (linear correlation between SHAP values), $\rho^{NDCG}$ (similarity of SHAP feature importance rankings),
$\rho^{Scores}$ (linear correlation between anomaly scores), $J$ (jaccard index between anomaly predictions) for each dataset. The mean matrices across all
datasets are presented in Figure~\ref{fig:matrices}, where hierarchical clustering was applied to optimize the ordering. Visually, two clusters of
models emerge. First, COPOD and ECOD consistently group, which is expected as both algorithms rely on similar distributional assumptions. Second, a larger
group comprising OCSVM, AutoEncoder, IForest, PCA, KNN, and CBLOF exhibits high correlation. These algorithms share underlying characteristics related to
distance metrics and data encoding strategies.

The Mantel test is used to assess whether the similarity matrices introduced above are statistically correlated. Table~\ref{tab:mantel} reports the mean Mantel
correlations across datasets. We observe a strong correlation ($r_M=0.83$) between $\rho^{PS}$ and $\rho^{NDCG}$, confirming that SHAP-based similarity is
consistent across feature importance magnitude and ranking. Similarly, $\rho^{Scores}$ and $J$ show a high correlation ($r_M=0.76$), which is intuitive since
models with similar score distributions tend to produce similar binary predictions. Finally, the correlation between $\rho^{PS}$ and $J$ ($r_M=0.67$) suggests
that detectors sharing similar reasoning patterns (as captured by SHAP) also tend to give similar anomaly predictions. However, the similarity between each
model depends on the dataset. Specifically, Table~\ref{tab:corrmatrices} shows the correlations between $\rho^{PS}$ and $\rho^{Scores}$, and between
$\rho^{PS}$ and $J$ for each dataset. Usually, both pairs are correlated. Some datasets show high correlations between matrices ($r_M > 0.5 $) like AN, PI, TH,
or VO. Some others have moderate or low correlation between the two matrices. Finally, some interesting datasets are BR, LY or YE where
$r_M(\rho^{PS},\rho^{Scores}) << r_M(\rho^{PS},J)$.

\begin{table}[htb!]
	\centering
	\caption{Mean Mantel correlations between distance matrices, averaged across datasets. All correlations are significant ($p \leq 0.004$).}
	\begin{tabular}{lcccc}
		\toprule
		                & $\rho^{PS}$ & $\rho^{NDCG}$ & $\rho^{Scores}$ & $J$  \\
		\midrule
		$\rho^{PS}$     & 1.00        & 0.83          & 0.55            & 0.67 \\
		$\rho^{NDCG}$   &             & 1.00          & 0.54            & 0.57 \\
		$\rho^{Scores}$ &             &               & 1.00            & 0.76 \\
		$J$             &             &               &                 & 1.00 \\
		\bottomrule
	\end{tabular}

	\label{tab:mantel}
	\vspace{5pt} 
\end{table}

\begin{table}[htb!]
	\centering
	\caption{Mantel correlations ($\times 10^2$) between distance matrices for each dataset.}
	\setlength{\tabcolsep}{5pt}

	\resizebox{\textwidth}{!}{
		\begin{tabular}{l*{16}{c}}
			\toprule
			Dataset                        & AN & BR & GL & HE & LY & MA & PB & PI & ST & TH & VE & VO & WB & WL & WN & YE \\
			\midrule
			$r_M(\rho^{PS},\rho^{Scores})$ & 75 & 22 & 42 & 26 & 36 & 56 & 33 & 55 & 56 & 72 & 26 & 56 & 18 & 49 & 26 & 38 \\
			$r_M(\rho^{PS},J)$             & 78 & 51 & 24 & 51 & 84 & 51 & 66 & 52 & 53 & 76 & -4 & 58 & 15 & 49 & 30 & 62 \\
			\bottomrule
		\end{tabular}
	}
	\label{tab:corrmatrices}
\end{table}

Overall, the similarity of models is correlated with their predictions. In the next sections, we investigate combining models that are far apart in the
similarity matrices to enhance the overall results of an ensemble.

\subsection{Aggregate Models Predictions}

The objective of an ensemble is to aggregate the predictions of multiple detectors to create a more robust method. A challenge in this kind of method is how to
aggregate each prediction. Each anomaly detector produces an anomaly score and a prediction indicating whether a data point is normal or anomalous. To truly
leverage each model's specialty, we utilize anomaly scores as input for our aggregation function. Known aggregation methods for these scores include using the
maximum, taking the average, or ranking the scores and averaging the ranks~\cite{aggarwal2016outlier}. We evaluated these functions on all possible ensembles
of 3 models from our 14-model pool (resulting in 364 ensembles) on each dataset. Given the class imbalance of each dataset, we rely on the Area Under the
Precision-Recall Curve (AUCPR). As shown in Table~\ref{tab:beststrat}, rank aggregation yields superior results on 11 out of 16 datasets. Consequently, we
adopt rank aggregation for the remainder of this study.

\begin{table}[htb!]
	\centering
	\caption{Average AUCPR ($\times 10^2$) between all of the 364 ensembles with the 3 main aggregation strategies. Best results are in bold.}
	\setlength{\tabcolsep}{5pt}

	\setlength{\aboverulesep}{0pt}
	\setlength{\belowrulesep}{0pt}
	\renewcommand{\arraystretch}{1.2}

	\resizebox{\textwidth}{!}{
		\begin{tabular}{lcccccccccccccccc||c}
			\toprule
			Dataset & AN          & BR          & GL          & HE          & LY           & MA          & PB          & PI          & ST          & TH          & VE          & VO          & WB          & WL         & WN          & YE          & mean        \\
			\midrule
			Rank    & \textbf{26} & \textbf{98} & 18          & \textbf{84} & \textbf{100} & \textbf{30} & \textbf{60} & \textbf{55} & \textbf{62} & \textbf{58} & 18          & \textbf{41} & \textbf{97} & 6          & 49          & 37          & \textbf{52} \\
			Max     & 21          & 59          & \textbf{22} & 64          & 68           & 8           & 35          & 45          & 35          & 28          & \textbf{28} & 11          & 41          & \textbf{9} & \textbf{53} & \textbf{40} & 35          \\
			Mean    & 9           & 49          & 19          & 47          & 36           & 4           & 12          & 42          & 22          & 5           & \textbf{28} & 8           & 25          & 7          & 31          & 39          & 24          \\
			\bottomrule
		\end{tabular}
	}
	\label{tab:beststrat}
\end{table}

\subsection{Complementarity}

To select the most dissimilar models, we computed the dissimilarity matrices $D^P_{\mathcal{D}}$ for each dataset $\mathcal{D}$, which gives matrices on how
far two detectors are for each dataset $\mathcal{D}$. From the pool of 14 models, we constructed ensembles of size $n=3$. Table~\ref{tab:corrf1dist} presents
the correlations between the AUCPR of ensembles and the diversity distances for each similarity matrix. A positive correlation between diversity and
performance implies that increasing ensemble diversity leads to superior detection results.

\begin{table}[htb!]
	\centering
	\caption{Correlation ($\times 10^2$) between the distance of models and the AUCPR of the ensembles. Best results are in bold.}

	\setlength{\tabcolsep}{5pt}

	\setlength{\aboverulesep}{0pt}
	\setlength{\belowrulesep}{0pt}
	\renewcommand{\arraystretch}{1.2}

	\resizebox{\textwidth}{!}{

		\begin{tabular}{lcccccccccccccccc}
			\toprule
			metric          & AN          & BR          & GL         & HE           & LY          & MA          & PB          & PI          & ST          & TH          & VE          & VO          & WB          & WL          & WN          & YE          \\
			\midrule
			$\rho^{PS}$     & -20         & \textbf{-0} & \textbf{5} & \textbf{-28} & 15          & \textbf{-9} & -18         & \textbf{37} & \textbf{40} & -38         & -44         & -66         & \textbf{51} & 55          & 43          & 29          \\
			$\rho^{NDCG}$   & \textbf{39} & -17         & -2         & -32          & -0          & -68         & \textbf{21} & 14          & 15          & \textbf{19} & -30         & -54         & 36          & \textbf{62} & \textbf{48} & 37          \\
			$\rho^{Scores}$ & -1          & -90         & -33        & -48          & -29         & -18         & -17         & 30          & 15          & -12         & -12         & \textbf{-0} & -11         & 46          & 7           & 34          \\
			$J$             & -32         & -59         & -37        & -31          & \textbf{16} & 3           & -4          & 24          & 36          & -48         & \textbf{-7} & -3          & -28         & 55          & 1           & \textbf{45} \\
			\bottomrule
		\end{tabular}
	}
	\label{tab:corrf1dist}
\end{table}

The table shows three key insights. Firstly, selecting diversity from SHAP values tends to give a higher diversity than model outputs (scores and predictions).
On 11 out of 16 datasets, using $\rho^{PS}$ and $\rho^{NDCG}$ as diversity, tends to give better results than using $\rho^{Scores}$ and $J$. Secondly, as stated previously,
diversity contributes to ensemble learning by expanding the range of detected anomalies. However, effective model selection should also account for individual
model performance, a factor not explicitly optimized in this study. This limitation likely explains why the correlations between distance metrics and
performance remain moderate. Furthermore, in some datasets, these correlations are consistently negative, suggesting that diversity is not always beneficial.
This phenomenon occurs when a single model dominates the others; in such cases, the primary condition for a successful ensemble is the inclusion of this
specific model. Additionally, negative correlations can arise in highly complex datasets where all models exhibit poor performance. In these scenarios, even
high diversity cannot compensate for the lack of meaningful detection, resulting in an ineffective ensemble. Finally, it is interesting to note that SHAP
values and scores do not show the same diversity as correlations vary between datasets. For example, on the dataset WB, $\rho^{Scores}$ and $J$ are
negative, while are positive $\rho^{PS}$ and $\rho^{NDCG}$. Consequently, the two metrics highlight different diversities.

\subsection{Diversity and Individual Performances}

While we have shown that diversity enhances ensemble performance by expanding anomaly coverage, our earlier analysis neglected individual model accuracy. Here,
we refine our results to demonstrate that despite the value of diversity, individual model quality remains a critical factor. For the purpose of this section, we use $\rho^{PS}$ to quantify diversity.

\begin{figure}[htb!]
	\centering
	\begin{subfigure}[b]{0.45\textwidth}
		\centering
		\includegraphics[width=\textwidth]{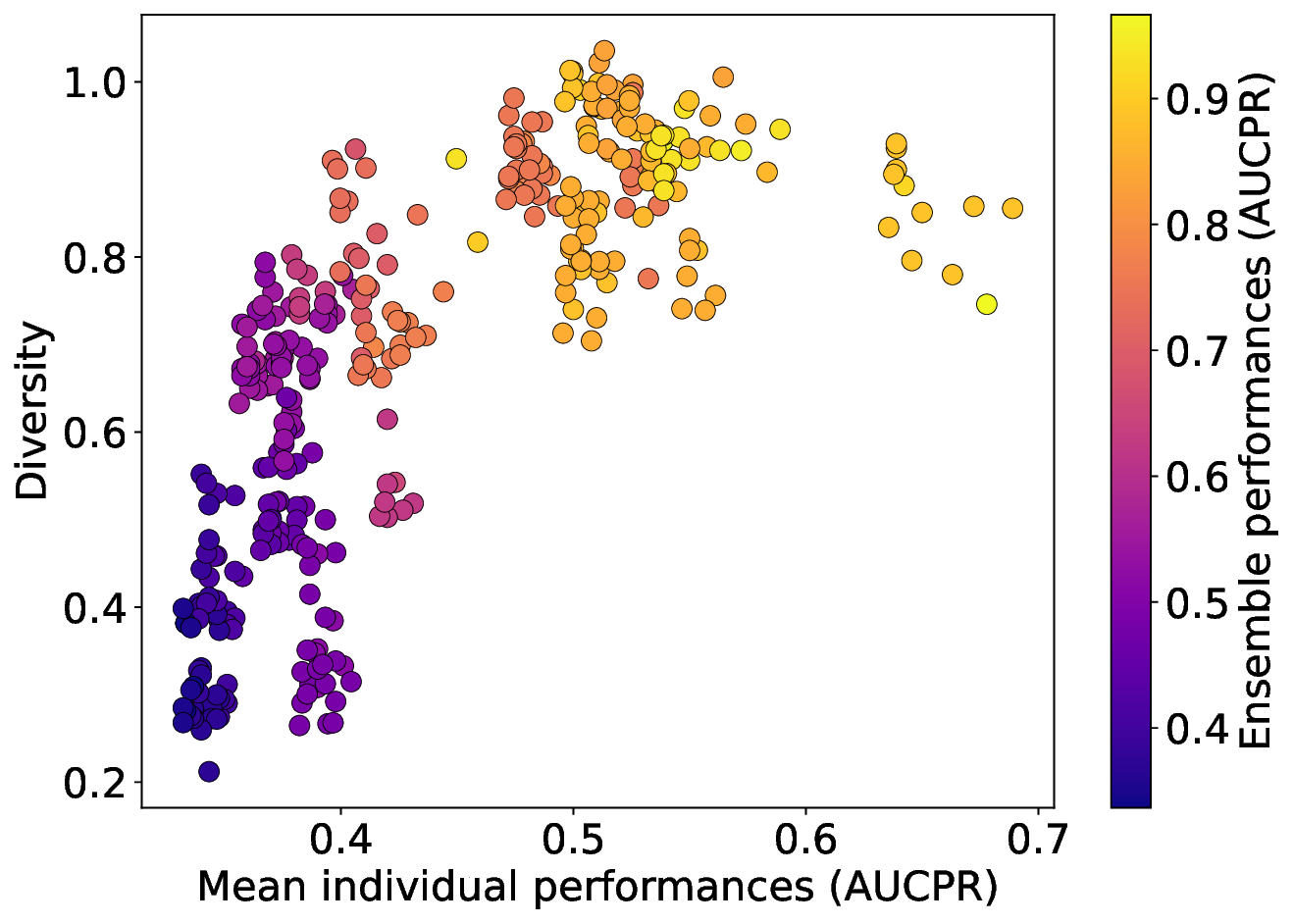}
		\caption{Dataset Lymmography (LY).}
		\label{fig:performances-a}
	\end{subfigure}
	\hfill
	\begin{subfigure}[b]{0.45\textwidth}
		\centering
		\includegraphics[width=\textwidth]{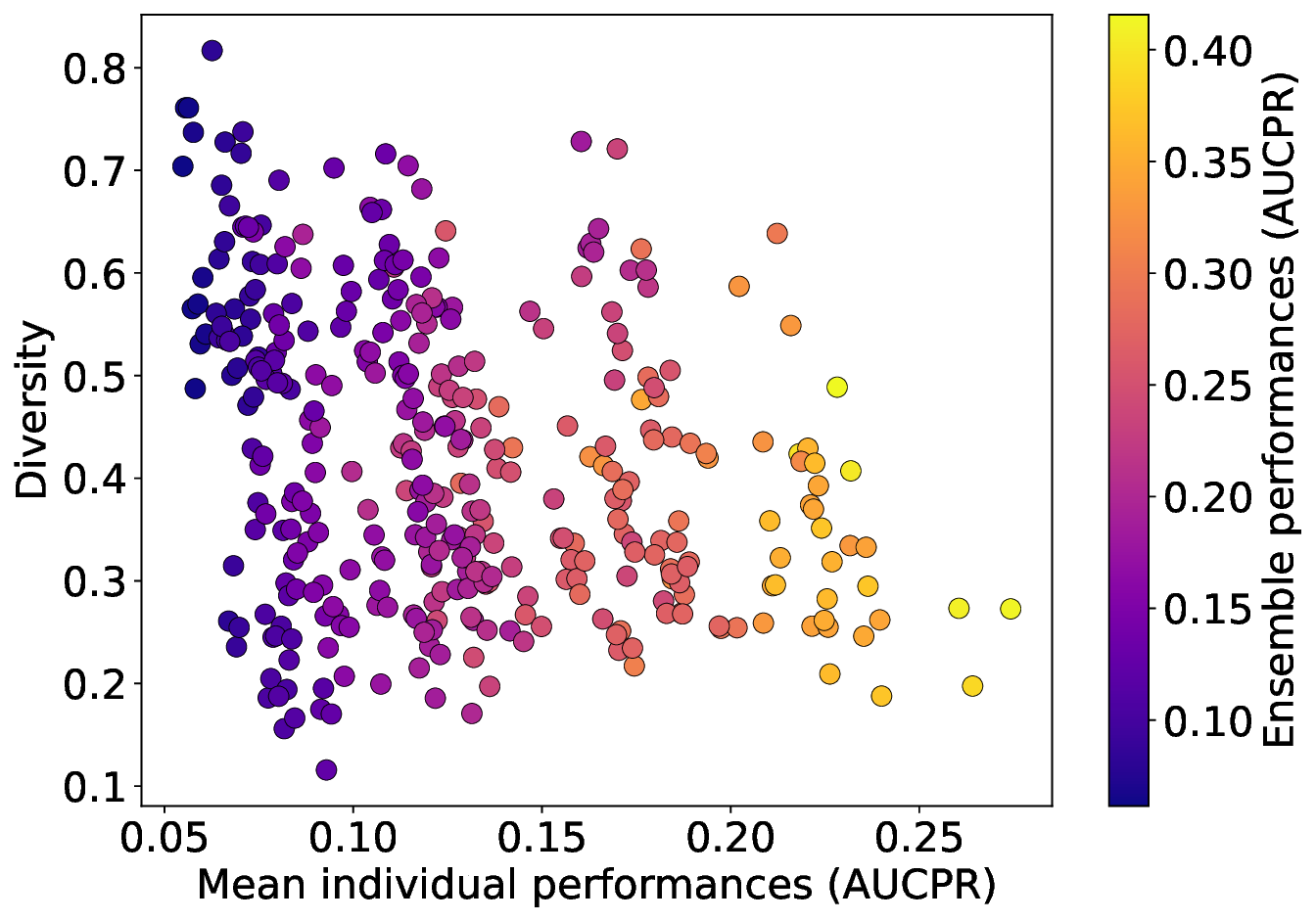}
		\caption{Dataset Vowels (VO).}
		\label{fig:performances-b}
	\end{subfigure}
	\caption{Relationship between ensemble diversity (given by $\rho^{PS}$) and average individual performance. Each point represents an ensemble of models. The color scale indicates the ensemble's overall performance (AUCPR).}
	\label{fig:performances}
\end{figure}

Figure~\ref{fig:performances} illustrates the importance of individual model performance for achieving an effective ensemble. Specifically, Figure
\ref{fig:performances-a} presents the results for the LY dataset. A clear correlation between diversity, individual performance, and ensemble accuracy is
observable on this dataset. However, this correlation is not always beneficial. In certain scenarios, enforcing diversity can be detrimental to the ensemble.
For instance, Figure~\ref{fig:performances-b} reveals that for the VO dataset, diversity offers negligible gain and can even lead to suboptimal results.

To assess the relative impact of model quality versus diversity, we performed a linear regression to predict the ensemble performance gain using standardized
mean individual performance and diversity scores. We confirmed the stability of these estimates by ruling out multicollinearity (see Appendix~\ref{sec:appendixb}). The resulting weights, presented in Table~\ref{tab:weightslr}, indicate that while individual performance is
generally the dominant factor, diversity plays a crucial complementary role. In 12 out of 16 datasets, the diversity coefficient is positive, confirming its
value as a performance booster. Notably, for the WB and LY datasets, the diversity weight rivals or even exceeds that of individual performance (Ratios of 1.2
and 0.8 respectively). However, overall, diversity plays a secondary yet valuable role in UAD ensemble design.

\begin{table}[t!]
	\centering
	\caption{Linear regression weights ($\times 10^2$) predicting ensemble performance from mean individual performance and diversity, along with their ratio (green: positive; red: negative).}
	\setlength{\tabcolsep}{5pt}

	\setlength{\aboverulesep}{0pt}
	\setlength{\belowrulesep}{0pt}
	\renewcommand{\arraystretch}{1.2}

	\resizebox{\textwidth}{!}{
		\begin{tabular}{lcccccccccccccccc}
			\toprule
			             & AN         & BR         & GL          & HE          & LY         & MA         & PB         & PI         & ST         & TH         & VE          & VO   & WB  & WL  & WN   & YE   \\
			\midrule
			Indiv. Perf. & 2.7        & 6.3        & 1.1         & 6.5         & 10.7       & 3.4        & 5.0        & 1.7        & 5.6        & 5.7        & 0.1         & 7.5  & 2.4 & 0.0 & 8.0  & 0.3  \\
			Diversity    & 0.5        & 2.0        & 0.1         & 0.5         & 8.4        & 0.7        & 2.2        & 0.2        & 0.3        & 1.7        & -0.0        & -0.6 & 2.8 & 0.0 & -0.1 & -0.0 \\
			\hline
			Ratio        & \cpos{0.2} & \cpos{0.3} & \cpos{0.1}  & \cpos{0.1}  & \cpos{0.8} & \cpos{0.2} & \cpos{0.4} & \cpos{0.1} & \cpos{0.0} & \cpos{0.3} & \cneg{-0.1} &
			\cneg{-0.1}  & \cpos{1.2} & \cpos{0.4} & \cneg{-0.0} & \cneg{-0.0}                                                                                                                              \\ \bottomrule
		\end{tabular}
	}
	\label{tab:weightslr}
\end{table}



\section{Conclusion}
\label{sec:conclusion}

We presented a methodology for selecting models in UAD ensembles based on the similarity of their explanations. We demonstrated that diversity can be usefuland enhance results, particularly on datasets where individual model performances are close to each other. We analyzed four different diversity metrics: two based on SHAP explanations and two based directly on model outputs. These metrics yielded different results for ensemble model selection, indicating that the diversity they highlight is different. Finally, we established that despite the benefits of diversity, individual model performance remains the deciding factor: weak but diverse models cannot outperform strong but similar models. Nevertheless, we believe that explainability should be more considered in UOMS as it provides new information on model behavior.

A limitation of our approach lies in the computational cost of SHAP, which can become prohibitive for datasets with a large number of instances or features. However, the computational cost can be alleviated by using approximation techniques or surrogate models. Additionally, our strategy is agnostic to the explanation method, allowing for the use of faster interpretability techniques if required. For instance, in~\cite{cavus2025beyond}, authors recently demonstrated how aggregating Partial Dependence Profiles (PDP) across a set of near-optimal models can provide reliable metrics for explanation uncertainty and robustness.

Furthermore, while we successfully model the diversity, our study did not address the optimization of individual models; we neither investigated the individual performances of each anomaly detector nor their best
hyperparameters.

In future work, we plan to investigate the divergence between SHAP-based and raw output similarities to refine the model selection process. Furthermore, since individual performance is critical, we plan to integrate methods to characterize model performance into our methodology. Finally, we aim to extend our strategy to time series UAD, where ensemble methods can be even more powerful, as the problem is even more complex~\cite{levy2025timeciel}.

\begin{credits}
	\subsubsection{\ackname} The research presented in this paper has received funding from the National Association for Research and Technology under grant number CIFRE 2023/1398 and from Soben company. The authors also thank the National Agency for Research for funding the project MIMICO under grant number ANR-24-CE23-0380.

	\subsubsection{\discintname} The authors declare no Conflict of interest.

\end{credits}

\section*{Appendix}

\section{Description of the datasets}
\label{sec:appendixa}
\begin{table}[H]
	\centering
	\caption{Description of the datasets used.}

	\begin{tabular}{llcccc}
		\toprule
		Code & Name         & \# Samples & \# Features & \# Anomaly (\%) & Category    \\
		\midrule
		AN   & annthyroid   & 7200       & 6           & 534 (7.42)      & Healthcare  \\
		BR   & breastw      & 683        & 9           & 239 (34.99)     & Healthcare  \\
		GL   & glass        & 214        & 7           & 9 (4.21)        & Forensic    \\
		HE   & Hepatitis    & 80         & 19          & 13 (16.25)      & Healthcare  \\
		LY   & Lymphography & 148        & 18          & 6 (4.05)        & Healthcare  \\
		MA   & mammography  & 11183      & 6           & 260 (2.32)      & Healthcare  \\
		PB   & PageBlocks   & 5393       & 10          & 510 (9.46)      & Document    \\
		PI   & Pima         & 768        & 8           & 268 (34.90)     & Healthcare  \\
		ST   & Stamps       & 340        & 9           & 31  (9.12)      & Document    \\
		TH   & thyroid      & 3772       & 6           & 93  (2.47)      & Healthcare  \\
		VE   & vertebral    & 240        & 6           & 30  (12.50)     & Biology     \\
		VO   & vowels       & 1456       & 12          & 50  (3.43)      & Linguistics \\
		WB   & WBC          & 223        & 9           & 10  (4.48)      & Healthcare  \\
		WL   & Wilt         & 4819       & 5           & 257 (5.33)      & Botany      \\
		WN   & wine         & 129        & 13          & 10  (7.75)      & Chemistry   \\
		YE   & yeast        & 1484       & 8           & 507 (34.16)     & Biology     \\
		\bottomrule
	\end{tabular}
	\label{tab:datasets}
\end{table}

\section{Assessment of Multicollinearity}
\label{sec:appendixb}

To ensure the reliability of the estimated coefficients in our linear regression model, we assessed the potential for multicollinearity between the predictor variables: average individual model performance and ensemble diversity. Multicollinearity can inflate the variance of coefficient estimates, making it difficult to isolate the individual influence of each predictor on the ensemble performance.

We evaluated this risk using the Variance Inflation Factor (VIF) across all examined datasets. The VIF for a prediction $i$ is calculated as
\begin{equation*}
	VIF_i = \frac{1}{1 - R_i^2},
\end{equation*}
with $R_i^2$ the coefficient of determination of a regression of predictor $i$ on all other predictors. Traditionally, a $VIF > 5$ or $10$ indicates problematic multicollinearity. As shown in Table \ref{tab:multicollinearity}, our results consistently fall well below these thresholds. As $VIF_X = VIF_Y$ for two predictors, we reported a single value.

\begin{table}[h!]
	\centering
	\caption{VIF values for each datasets}

	\setlength{\tabcolsep}{5pt}
	\resizebox{\textwidth}{!}{
		\begin{tabular}{l*{16}{c}}
			\toprule
			Dataset & AN  & BR  & GL  & HE  & LY  & MA  & PB  & PI  & ST  & TH  & VE  & VO  & WB  & WL  & WN  & YE  \\
			\midrule
			VIF     & 1.0 & 1.3 & 1.1 & 1.7 & 2.3 & 1.1 & 1.0 & 1.1 & 1.3 & 1.3 & 1.0 & 1.1 & 1.2 & 2.0 & 1.1 & 1.5 \\
			\bottomrule
		\end{tabular}
	}
	\label{tab:multicollinearity}
\end{table}

These results confirm the predictors are distinct, ensuring stable and interpretable estimates of their individual influences.

\printbibliography

\end{document}